\documentclass{article}
\usepackage{spconf,amsmath,graphicx}
\usepackage{amssymb,booktabs,makecell,xcolor}
\usepackage{hyperref}

\makeatletter
\let\spikeposeOriginalBibliography\thebibliography
\renewcommand{\thebibliography}[1]{%
  \spikeposeOriginalBibliography{#1}%
  \setlength{\itemsep}{-2pt}%
}
\makeatother

\definecolor{spblue}{HTML}{2F6DA8}

\newcommand{\newresult}[1]{#1}
\newcommand{\PCKHB}{\mathrm{PCK}_{\mathrm{HB}}}

\title{SpikePose: Lightweight Video Human Pose Estimation}

\name{Yunhao Huang, Kaushik Bhargav Sivangi, Tianyi Yu, Fani Deligianni}
\address{School of Computing Science, University of Glasgow, UK}

\makeatletter
\renewcommand{\@maketitle}{%
  \newpage
  \null
  \vskip 0em
  \noindent\begin{minipage}{\textwidth}
  \footnotesize\itshape
  This work has been submitted to the IEEE for possible publication. Copyright may be transferred without notice, after which this version may no longer be accessible.
  \end{minipage}\par
  \vskip 0.65em
  \begin{center}
    {\large \bf \@title \par}
    \vskip 1.5em
    {\large \lineskip .5em
      \begin{tabular}[t]{c}\@name \\ \@address
      \end{tabular}\par}
  \end{center}
  \par
  \vskip 1.5em
}
\makeatother

\begin{document}
\ninept 
\maketitle

\begin{abstract}

Efficient video human pose estimation must balance localisation accuracy, motion quality, and energy cost. We present SpikePose, a lightweight video pose estimator that pairs a frame-reset, single-step spiking encoder with motion-aligned heatmap memory (MAM). The encoder supplies per-frame spatial evidence, while MAM estimates joint displacement, aligns historical heatmaps, and adaptively fuses them with current predictions. By confining temporal reasoning to a compact, lightweight module outside the spiking backbone, SpikePose helps bridge the gap between spiking and classical Artificial Neural Networks (ANN) on large-scale, non-event-driven video tasks. We evaluate spatial accuracy, motion errors, and estimated arithmetic energy on MPII and a fixed NTU60 Cross-Subject subset. Against the architecture-matched ANN with MAM, SpikePose cuts estimated arithmetic energy by 91.4\% at a cost of just 1.12  points in head–bone-normalised PCK at 0.5 threshold ($\PCKHB(0.5)$). Within SpikePose, adding MAM reduces velocity and acceleration errors by 28.13\% and 31.46\% while slightly improving localisation, at a cost of only 0.003M parameters and 0.0023 analytical GFLOPs per frame. Code will be released on GitHub upon acceptance.

\end{abstract}

\begin{keywords}
spiking neural networks, video human pose estimation, single-step spiking inference, heatmap memory, efficient inference
\end{keywords}

\section{Introduction}

2D Video human pose estimation (HPE) requires accurate joint localisation and reliable motion trajectories. Frame-wise backbones such as ResNet-50 Baselines~\cite{he2016resnet,xiao2018simple} and HRNet~\cite{sun2019hrnet} provide strong spatial representations but do not explicitly use video history. Temporal approaches introduce feature warping, aggregation, and space--time modelling~\cite{pfister2015flowing,bertasius2019posewarper,liu2022famipose,he2024dsta,feng2025glsmamba}. Processing high-dimensional temporal features increases computational costs, motivating a joint examination of localisation, motion quality, and arithmetic energy efficiency.

Spiking neural networks (SNNs) offers an alternative computational primitive for spatial encoding~\cite{neftci2019surrogate}. Bio-inspired SNN units accumulate inputs into a continuous membrane potential and emit a discrete spike after crossing a threshold before the potential resets or decays. This thresholded, event-driven behaviour replaces the dense multiplication operations of analog neural networks with sparse, accumulation-based updates, which is a key reason of the energy efficiency broadly reported for SNNs. SNNs are, however, natively formulated for asynchronous event streams. A central open challenge is scaling them to tasks such as image and video understanding that are not intrinsically event-based, where inputs arrive as dense synchronous frames~\cite{kudithipudi2025neuromorphic}. Thresholded spiking activations have been applied to object detection and visual backbones by supplying static images and unrolling membrane dynamics over several steps~\cite{kim2020spikingyolo,yao2023spikedriven}. Video introduces an important distinction: the internal update index $\tau$ describes neural dynamics applied to an observation, whereas the physical frame index $t$ advances only when a new observation is acquired. Increasing $\tau$ at a fixed $t$ may refine the same image representation, but it cannot reveal inter-frame displacement. Directly retaining an unaligned membrane state is also problematic because a moving joint may occupy a different location in the next image. SNN recurrence is therefore neither cost-free nor inherently motion-aware video memory.

SpikePose assigns spatial encoding to a frame-reset, single-step SNN and cross-frame reasoning to motion-aligned heatmap memory (MAM). MAM estimates joint displacement from consecutive predictions and aligns historical heatmaps before fusion. Related motion-aware heatmap regression incorporates joint movement into heatmap targets~\cite{song2024motionaware}. Here, motion instead controls a recurrent heatmap state. This design keeps temporal processing outside the high-dimensional backbone. Our contributions are threefold:

\begin{itemize}
  \item A novel compact 2D video HPE system that combines single-step spiking spatial encoding with an independent temporal state, evaluated against an architecture-matched ANN to quantify its accuracy--energy trade-off.
  \item A lightweight MAM that aligns and adaptively retains heatmap history, to reduce frame-wise SNN's motion errors with minimal computational overheads.
  \item Comprehensive validation on both the MPII image benchmark and the NTU60 video dataset, with component and filtering analyses assessing individual architectural contribution.

\end{itemize}

\begin{figure*}[!t]
\centering
\includegraphics[width=0.82\textwidth]{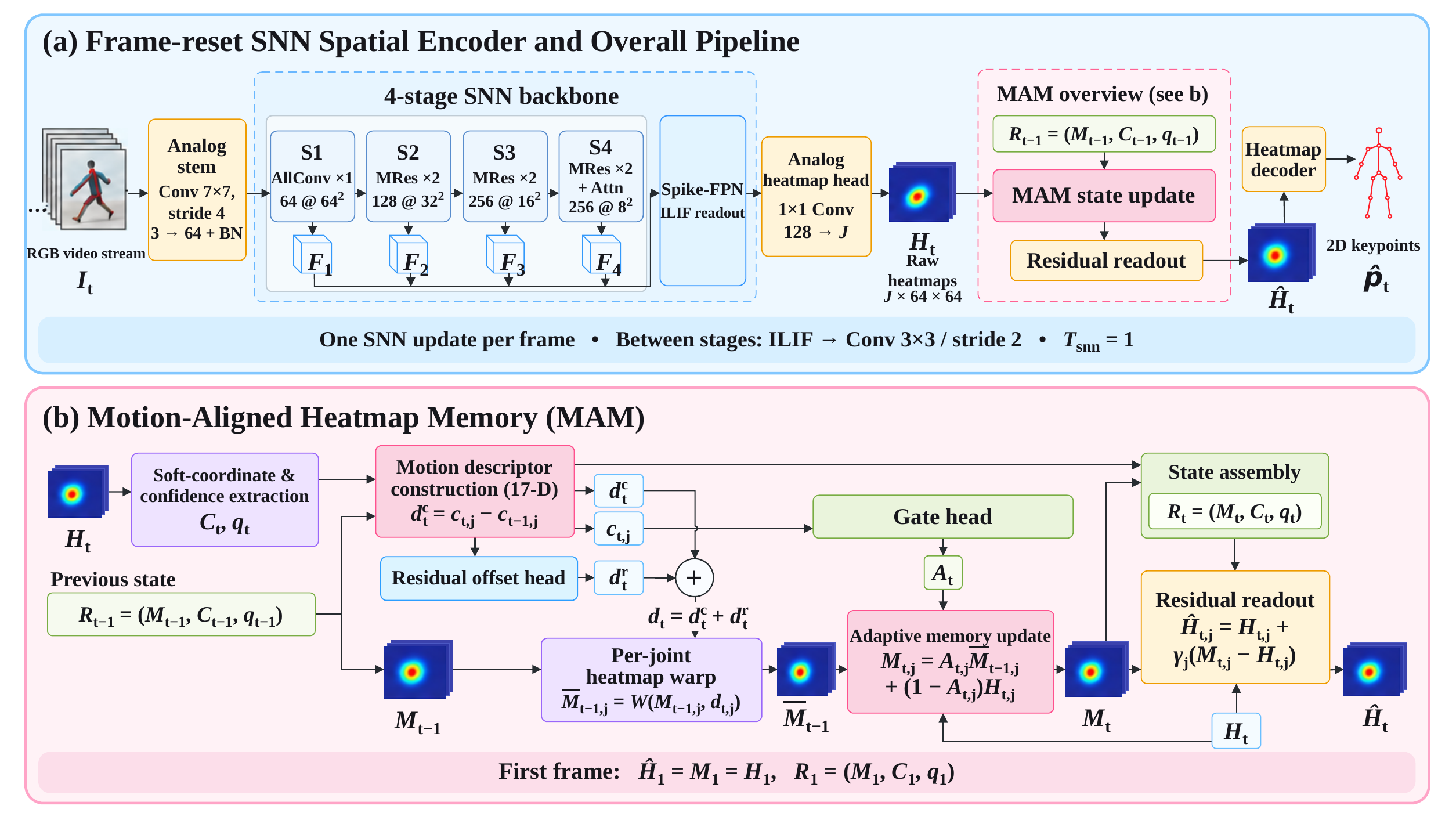}
\caption{Overview of the proposed SpikePose pipeline. The frame-reset spiking encoder extracts per-frame spatial evidence, while MAM performs compact cross-frame motion estimation, heatmap alignment, memory update, and residual decoding.}
\label{fig:overview}
\end{figure*}

\section{Methods}
\label{sec:method}


Figure~\ref{fig:overview} shows the overall pipeline. Throughout this paper, SpikePose denotes the complete encoder--MAM system, whereas SpikePose w/o MAM denotes its frame-wise SNN variant without MAM. Specifically, the \textit{frame-reset spiking spatial encoder} ($f_\Theta^{\text{SNN}}$) processes each RGB frame independently using a single-step spiking inference schedule ($T_{\text{snn}}=1$). By resetting the internal neuronal state before every frame, it extracts per-frame spatial heatmaps ($H_t$) without carrying unaligned membrane history across frames. To handle cross-frame reasoning, the lightweight \textit{motion-aligned heatmap memory (MAM) module} ($g_\Phi$) maintains a compact state $R_t=(M_t,C_t,q_t)$ across frames. Operating entirely outside the high-dimensional backbone, MAM estimates joint displacement, warps historical heatmaps, and applies adaptive gating and residual fusion to significantly reduce motion errors with minimal computational overhead.

\subsection{Single-Step Frame-Reset Spiking Spatial Encoder}

The encoder maps a $256\times256$ crop to 16 joint heatmaps of size $64\times64$ through four progressively downsampling stages. An analog stem first extracts low-level features before the representation is handed off to spiking stages, which use I-LIF--Conv--BN operations to perform downsampling while keeping activations in the integer, event-driven spiking domain. These stages further rely on membrane-shortcut blocks~\cite{hu2024msresnet} to preserve gradient flow and maintain training stability despite the network's discrete, threshold-based activations. The final stage incorporates spatial attention to help the network focus on informative joint regions, while a spiking feature pyramid~\cite{lin2017fpn} fuses multi-scale features into a unified representation before a lightweight analog head decodes the final heatmaps. This design keeps almost all computation within the energy-efficient spiking domain, reserving analog operations for the input and output stages where precision is most critical.

For internal update $\tau$, the integer-valued LIF neuron~\cite{luo2024ilif} follows
\begin{align}
U_{\tau}
&=\begin{cases}
X_{\tau}, & \tau=1,\\
\lambda(U_{\tau-1}-\vartheta S_{\tau-1})+X_{\tau}, & \tau>1,
\end{cases}\\
S_{\tau}
&=\operatorname{round}\!\left[
\operatorname{clip}\!\left(U_{\tau}/\vartheta,0,K\right)\right],
\end{align}
We use $\lambda=0.9$, $\vartheta=1$, and $K=4$, with a straight-through derivative inside the unclipped interval. The selected model uses only $\tau=1$: its RGB input is supplied once, subsequent spiking convolutions consume integer activations, and membrane shortcuts remain real-valued. Here, $K$ bounds spike magnitude rather than update count. Larger update counts are diagnostic controls in Sec.~\ref{sec:experiments}.

\subsection{Lightweight Motion-Aligned Heatmap Memory}

MAM maintains a compact heatmap-domain state through motion alignment, gated retention, and residual fusion. For joint $j$, let $\mathbf x$ index heatmap pixels.

Here, $\mathbf C_t=[\mathbf c_{t,1},\ldots,\mathbf c_{t,J}]^\top$ and $\mathbf q_t=(q_{t,1},\ldots,q_{t,J})^\top$, with $J=16$. The soft coordinate and confidence are
\begin{align}
P_{t,j}(\mathbf x) &= \operatorname{softmax}_{\mathbf x}(H_{t,j})(\mathbf x),\\
\mathbf c_{t,j} &= \sum_{\mathbf x}P_{t,j}(\mathbf x)\mathbf x, & q_{t,j} &= \max_{\mathbf x}P_{t,j}(\mathbf x).
\end{align}
Coordinates and confidences are extracted only from raw spatial heatmaps, not from memory or fused outputs. The coarse inter-frame displacement is given as:

\begin{equation}
\mathbf d^c_{t,j} = \mathbf c_{t,j}-\mathbf c_{t-1,j}.
\end{equation}

The 17-D descriptor $\mathbf z_{t,j}$ concatenates the current and previous coordinates and confidences, $\mathbf d^c_{t,j}$ and its magnitude, and an 8-D joint embedding.

Directly averaging $M_{t-1,j}$ and $H_{t,j}$ can blur spatially displaced peaks. A two-layer offset head therefore predicts a bounded residual correction, after which bilinear sampling aligns the previous memory:
\begin{align}
\mathbf d_{t,j} &= \mathbf d^c_{t,j} + 2\tanh\!\left(h_{\mathrm{off}}(\mathbf z_{t,j})\right),\\
 \overline M_{t-1,j} &= \mathcal W(M_{t-1,j},\mathbf d_{t,j}).
\end{align}

The residual is limited to two heatmap pixels per axis, equivalent to eight pixels at the input resolution, so it locally refines rather than replaces the coordinate displacement. The warp $\mathcal W$ uses bilinear sampling with boundary padding.

A lightweight gating head predicts a joint-wise history-retention gate shared across spatial locations:
\begin{equation}
A_{t,j} = 0.02 + 0.93\,\sigma\!\left(h_{\mathrm{gate}}(\mathbf z_{t,j})\right).
\end{equation}
The aligned memory and refined output are
\begin{align}
M_{t,j} &= A_{t,j}\overline M_{t-1,j} + (1-A_{t,j})H_{t,j},\\
\widehat H_{t,j} &= H_{t,j} + \gamma_j(M_{t,j}-H_{t,j}).
\end{align}
At $t=1$, MAM is bypassed: $\widehat H_1=M_1=H_1$; $\mathbf C_1$ and $\mathbf q_1$ are computed from $H_1$, and recurrent updates start at $t=2$. The learned per-joint coefficient $\gamma_j$ is initialised to zero and clipped to $[0,1]$, preserving the direct path at initialisation and bounding the memory correction.

\subsection{Training Objective}

Heatmap targets are unit-peak Gaussians with $\sigma=2$ heatmap pixels. Training back-propagates through each 16-frame window without detaching MAM state. The total objective combines the primary pose loss ($L_{pose}$) with auxiliary displacement supervision ($L_{offset}$). 
Specifically, $L_{pose}$ measures the mean-squared error on visible joints of the fused output $\widehat H_t$, normalised by the number of valid heatmap elements. $L_{offset}$ applies a Smooth $L1$ penalty to train the displacement branch using ground-truth coordinate differences across valid adjacent-frame pairs in $\mathcal V$:

\begin{equation}
\begin{aligned}
\mathcal L&=\mathcal L_{\mathrm{pose}}+0.01\mathcal L_{\mathrm{offset}},\\
\mathcal L_{\mathrm{offset}}
&=\underset{(t,j)\in\mathcal V}{\operatorname{mean}}
\operatorname{Smooth_{L1}}\!\left(
\mathbf d_{t,j},\mathbf c^*_{t,j}-\mathbf c^*_{t-1,j}\right).
\end{aligned}
\label{eq:loss}
\end{equation}
Here, $\mathbf c^*$ is the ground-truth coordinate in heatmap space, and $Smooth_{L1}$ is averaged over valid coordinate elements. The raw heatmaps have no additional pose loss. Displacement supervision trains the offset branch, while no velocity or acceleration loss is imposed on the final trajectories.

\section{Experiments}
\label{sec:experiments}

\textbf{Implementation details.}
 For MAM training and validation, each source video contributes two deterministic, non-overlapping 16-frame clips sampled at uniformly spaced temporal centres. Frames are consecutive (stride one), training shuffles only the window order, and all frames within a window share the same crop and augmentation. Training resets MAM state at each sampled clip boundary. Complete-video evaluation retains MAM state across inference chunks and resets it at video boundaries. The neuronal state is reset before every physical frame.

\subsection{Datasets and Evaluation Metrics}
\label{subsec:Datasets and Evaluation Metrics}

\noindent\textbf{Image dataset.}
MPII is an in-the-wild 2D HPE benchmark with about 25,000 images and over 40,000 annotated people covering diverse human activities~\cite{andriluka2014mpii}. We use its official validation split to evaluate the frame-wise spatial encoder.

\noindent\textbf{Video dataset.}
NTU RGB+D 60 (NTU60) is a large-scale indoor multimodal dataset containing 56,880 video samples across 60 action classes, with synchronised RGB, depth, infrared, and 3D skeleton streams~\cite{shahroudy2016ntu}. We evaluate the complete video model under the official Cross-Subject split using a frozen stratified test subset of 4,563 videos and 419,968 physical frames. Kinect RGB-plane joints are mapped from NTU-25 to the MPII-style 16-joint layout, and all methods use one fixed skeleton-derived $256\times256$ tube crop per complete video. Full videos are processed sequentially. The MAM state is carried across inference chunks and reset only at source-video boundaries. Duplicate spatial predictions from an overlapping terminal extraction window are discarded.

\noindent\textbf{Metrics.}
For MPII, we report the standard $\mathrm{PCKh}(0.5)$ metric~\cite{andriluka2014mpii}. For NTU60, following the same head-normalised PCK principle, we define a head--bone-normalised variant ($\PCKHB$). For joint $j$ in physical frame $t$,
\begin{align}
e_{tj}^{\mathrm{HB}}
&=
\frac{\lVert\hat{\mathbf p}_{tj}-\mathbf p_{tj}\rVert_2}
{\lVert\mathbf p_{t,\mathrm{head}}
-\mathbf p_{t,\mathrm{neck}}\rVert_2},
\\
\PCKHB(\alpha)
&=
\frac{\sum_{t,j}m_{tj}\,
\mathbb{I}[e_{tj}^{\mathrm{HB}}\leq\alpha]}
{\sum_{t,j}m_{tj}}.
\end{align}
Here, $m_{tj}=1$ only when the evaluated joint, head, and neck are valid and the head--neck bone length is non-zero. We report $\PCKHB(0.5)$ over all evaluated frames.

Mean Per-Joint Velocity Error (MPJVE) and Mean Per-Joint Acceleration Error (MPJAccE) measure the mean Euclidean discrepancies between the first- and second-order finite differences of predicted and ground-truth 2D joint trajectories, respectively~\cite{pavllo2019videopose3d,tchenegnon2022newspatiotemporallossfunction}. Their units are pixels/frame and pixels/frame$^2$. Both metrics are computed from reconstructed full-video trajectories, include valid differences across inference-window boundaries, and are averaged equally across videos.

\noindent\textbf{Computation and arithmetic energy.}
Costs are reported for $256\times256$ input crops. Analytical GFLOPs count dense convolutional/linear operations and supported additional operators, with two FLOPs per multiply-Accumulation operations(MACs). Here, it covers real-valued convolutional/linear paths. We distinguish MACs from spike-driven Accumulate operations(ACs) using integer-spike accounting \cite{yao2023spikedriven,luo2024ilif}. SNN activity is averaged over 28,404 16-frame windows from full-video evaluation. Using 45-nm FP32 reference costs of 4.6 pJ/MAC and 0.9 pJ/AC \cite{horowitz2014energy}, we estimate:
\begin{equation}
E_{\mathrm{proxy}}(\mathrm{mJ})
=\frac{4.6N_{\mathrm{MAC}}+0.9N_{\mathrm{AC}}}{10^9}.
\label{eq:energy}
\end{equation}
Softmax, warping, membrane updates, element-wise fusion, and memory/data movement are not fully covered. This energy proxy reflects analytical computational costs rather than directly measured device energy or execution latency.

\subsection{Main Results}
Table~\ref{tab:fullvideo} compares the complete SpikePose model with ANN and SNN baselines on the frozen NTU60 subset. We denote the architecture-matched ANN baseline without and with MAM as SpikePose--ANN w/o MAM and SpikePose--ANN, respectively.

\begin{table*}[!t]
\caption{Results on the frozen NTU60 Cross-Subject test subset. All rows use a complete video and MAM state reset only at video boundaries, as described in Sec.~\ref{sec:experiments}. MPJVE and MPJAccE average videos equally. GFLOPs and $E_{\mathrm{proxy}}$ are normalized per processed frame; parameters are in millions. Bold marks the best valid ANN and SNN result separately.}
\label{tab:fullvideo}
\centering
\small

\setlength{\tabcolsep}{3.0pt}
\begin{tabular*}{\textwidth}{@{\extracolsep{\fill}}lccccccc@{}}
\toprule
Method & Type & $\PCKHB(0.5)\uparrow$ & MPJVE$\downarrow$ & MPJAccE$\downarrow$ & GFLOPs/frame$\downarrow$ & $E_{\mathrm{proxy}}$ (mJ/frame)$\downarrow$ & Params (M)$\downarrow$ \\
\midrule
ResNet-50 & ANN & \newresult{85.81} & \newresult{3.79} & \newresult{6.06} & 14.555 & 33.277 & 34.000 \\
HRNet-W32 & ANN & \newresult{\textbf{86.35}} & \newresult{3.67} & \newresult{5.87} & 20.528 & 46.889 & 28.540 \\
SpikePose--ANN w/o MAM & ANN & \newresult{84.92} & \newresult{3.11} & \newresult{4.81} & \textbf{8.443} & \textbf{19.309} & \textbf{7.518} \\
SpikePose--ANN & ANN & \newresult{84.93} & \newresult{\textbf{3.03}} & \newresult{\textbf{4.69}} & 8.445 & 19.310 & 7.523 \\
\midrule
SpikeYOLO--Frame & SNN & \newresult{\textbf{84.73}} & \newresult{4.88} & \newresult{8.06} & 17.097 & 6.628 & 16.739 \\
SpikePose w/o MAM (Ours) & SNN & \newresult{83.63} & \newresult{4.87} & \newresult{8.02} & \textbf{8.459} & \textbf{1.627} & \textbf{7.520} \\
SpikePose (Ours) & SNN & \newresult{83.81} & \newresult{\textbf{3.50}} & \newresult{\textbf{5.50}} & 8.462 & 1.665 & 7.523 \\
\bottomrule
\end{tabular*}

\end{table*}


SpikePose achieves $\PCKHB(0.5)=83.81$, a MPJVE of 3.50, and an MPJAccE of 5.50. Relative to SpikePose w/o MAM, it increases $\PCKHB(0.5)$ by $0.18$ points while reducing MPJVE and MPJAccE by 28.13\% and 31.46\%, respectively. MAM adds 0.003M parameters and approximately 0.0023 analytical GFLOPs per processed frame. Overall, its main benefit is improved motion quality along with a subtle localisation gain. 

SpikePose bridges the performance gap that traditionally hinders spiking neural networks in complex video tasks, achieving competitive accuracy relative to its analog counterparts.
While, HRNet-W32 and the architecture-matched SpikePose--ANN exceed SpikePose in $\PCKHB(0.5)$ by 2.54 and 1.12 points, respectively (86.35 and 84.93 versus 83.81), SpikePose matches the operation count of the architecture-matched ANN while uses 58.78\% fewer analytical GFLOPs and 73.64\% fewer parameters than HRNet-W32. 
By successfully narrowing this accuracy deficit under a spike-driven framework, SpikePose establishes a scalable and highly efficient accuracy-motion-computation operating point for real-world deployment.

\subsection{Ablation Study}

\noindent\textbf{MPII spatial encoder results.}
Table~\ref{tab:spatial} compares spatial accuracy and estimated arithmetic energy on MPII. SpikePose--ANN w/o MAM and single-step SpikePose w/o MAM obtain $\mathrm{PCKh}(0.5)$ scores of 87.05 and 85.84, respectively. With similar analytical computation of 8.44 and 8.46 GFLOPs per crop, the single-step SNN reduces estimated arithmetic energy from 19.309 to 1.820 mJ per crop (90.6\%). Increasing $T_{\mathrm{snn}}$ to 2 improves PCKh to 86.16 but raises estimated energy to 2.984 mJ per crop.

\begin{table}[!t]
\caption{MPII validation comparison. PCKh uses flip testing. GFLOPs and $E_{\mathrm{proxy}}$ are per $256\times256$ person crop, including all $T_{\mathrm{snn}}$ updates and excluding flip testing. All arithmetic-energy estimates use the same MAC/AC accounting convention. PCK: PCKh $(0.5)$; Parameters are in millions.}
\label{tab:spatial}
\centering
\small

\setlength{\tabcolsep}{1.2pt}
\begin{tabular*}{\columnwidth}{@{\extracolsep{\fill}}lccccr@{}}
\toprule
Method & Model & \shortstack{PCKh$\uparrow$} & GFLOPs$\downarrow$ & \shortstack{$E_{\mathrm{proxy}}\downarrow$} & Params$\downarrow$ \\
\midrule
\makecell[cl]{SpikePose--ANN\\w/o MAM} & ANN & \newresult87.05 & \newresult\textbf{8.44} & \textbf{19.309} & \textbf{\newresult7.518} \\
ResNet-50~\cite{xiao2018simple} & ANN & 88.54 & 14.56 & 33.277 & 33.999 \\
HRNet-W32~\cite{sun2019hrnet} & ANN & \textbf{90.32} & 20.53 & 46.889 & 28.536 \\
\midrule
SpikeYOLO~\cite{luo2024ilif} & SNN & \textbf{87.79} & 17.10 & 6.579 & 16.739 \\
\makecell[cl]{SpikePose\\w/o MAM ($T_{\mathrm{snn}}=1$)} & SNN & 85.84 & \textbf{8.46} & \textbf{1.820} & \textbf{7.518} \\
\makecell[cl]{SpikePose\\w/o MAM ($T_{\mathrm{snn}}=2$)} & SNN & 86.16 & 16.92 & 2.984 & \textbf{7.518} \\
\bottomrule
\end{tabular*}

\end{table}

\begin{table}[!t]
\caption{Frame/update controls for the SpikePose spatial encoder on NTU60.
$V_mU_n$ uses $m$ distinct physical frames in $n$ SNN updates.
Frames are repeated uniformly when $n>m$.}
\label{tab:update-frame-controls}
\centering
\small

\setlength{\tabcolsep}{1.2pt}
\begin{tabular*}{\columnwidth}{@{\extracolsep{\fill}}ccccc@{}}
\toprule
Schedule & \shortstack{$\PCKHB(0.5)\uparrow$} & MPJVE$\downarrow$ & MPJAccE$\downarrow$ & \shortstack{$E_{\mathrm{proxy}}\downarrow$} \\
\midrule
$V_1U_1$ & \newresult{85.08}          & \newresult{5.30}          & \newresult{8.81}          & \textbf{1.788} \\
$V_1U_2$ & \newresult{\textbf{85.09}} & \newresult{5.20}          & \newresult{8.60}          & 2.982 \\
$V_1U_4$ & \newresult{84.75}          & \newresult{4.94}          & \newresult{8.13}          & \newresult{5.616} \\
$V_2U_2$ & \newresult{83.03}          & \newresult{5.32}          & \newresult{8.73}          & 2.969 \\
$V_2U_4$ & \newresult{83.14}          & \newresult{\textbf{4.89}} & \newresult{\textbf{7.95}} & 5.564 \\
$V_4U_4$ & \newresult{80.66}          & \newresult{6.22}          & \newresult{9.74}          & \newresult{5.602} \\
\bottomrule
\end{tabular*}

\end{table}

\noindent\textbf{Number of SNN updates.}
Table~\ref{tab:update-frame-controls} distinguishes repeated processing of one image from access to distinct physical frames. It shows the highest $\PCKHB(0.5)$ for $V_1U_2$ (85.09) and the lowest MPJAccE for $V_2U_4$ (7.95).

\begin{table}[!t]
\caption{Core MAM-component ablation on the frozen NTU60 test subset with state reset only at source-video boundaries.}
\label{tab:mam-ablation}
\centering
\small

\setlength{\tabcolsep}{1.0pt}
\begin{tabular*}{\columnwidth}{@{\extracolsep{\fill}}ccccc@{}}
\toprule
\multicolumn{3}{c}{MAM components} & \multicolumn{2}{c}{NTU60 test subset} \\
\cmidrule(lr){1-3}\cmidrule(l){4-5}
\shortstack{Motion\\alignment} & \shortstack{Adaptive\\gate} & \shortstack{Residual\\fusion} & $\PCKHB(0.5)\uparrow$ & MPJAccE$\downarrow$ \\
\midrule
$\times$ & $\times$ & $\times$ & 83.63 & 8.02 \\
$\times$ & $\checkmark$ & $\checkmark$ & 83.77 & 5.86 \\
$\checkmark$ & $\times$ & $\checkmark$ & 83.75 & 5.66 \\
$\checkmark$ & $\checkmark$ & $\times$ & \textbf{83.85} & 5.53 \\
$\checkmark$ & $\checkmark$ & $\checkmark$ & 83.81 & \textbf{5.50} \\
\bottomrule
\end{tabular*}

\end{table}

\noindent\textbf{MAM components.}
Table~\ref{tab:mam-ablation} compares separately trained configurations with spatial adaptation. Removing alignment disables both warping and offset supervision. Removing adaptive gating retains learned, input-independent per-joint retention, initialised to 0.08. Removing residual fusion outputs memory directly, equivalent to $\gamma_j=1$. The no-alignment and input-independent-retention variants obtain MPJAccE of 5.86 and 5.66, versus 5.50 for the full model. Without residual fusion, MPJAccE is 5.53 and $\PCKHB(0.5)$ is slightly higher. These results favor the full configuration for motion error, with only a small difference from direct memory output.

\begin{table}[!t]
\centering
\caption{Filtering comparison on NTU60. Base is SpikePose without MAM; + rows apply coordinate filters to Base. PCK: $\PCKHB(0.5)$; VE: MPJVE; AE: MPJAccE. L/M/H: low/medium/high motion-speed strata (1,521 videos each). EMA: exponential moving average~\cite{brown1961exponential}; One Euro: One Euro filter~\cite{spikepose_oneeuro2012}; SG: Savitzky--Golay filter~\cite{savitzky1964smoothing}.}
\label{tab:speed-strata}
\small

\setlength{\tabcolsep}{1.5pt}
\begin{tabular*}{\columnwidth}{@{\extracolsep{\fill}}lrrrrrr@{}}
\toprule
& \multicolumn{3}{c}{Overall}
& \multicolumn{3}{c}{AE by speed} \\
\cmidrule(lr){2-4}\cmidrule(l){5-7} Method & PCK$\uparrow$ & VE$\downarrow$ & AE$\downarrow$ & L$\downarrow$ & M$\downarrow$ & H$\downarrow$ \\
\midrule
Base       & 83.63 & 4.87 & 8.02 & 4.81 & 6.77 & 12.48 \\
+ EMA      & 83.69 & 4.15 & 6.60 & 3.70 & 5.51 & 10.59 \\
+ One Euro & 83.78 & 3.61 & 5.64 & \textbf{2.55} & \textbf{4.46} & 9.90 \\
+ SG       & 83.60 & 4.47 & 6.91 & 3.91 & 5.77 & 11.03 \\
SpikePose  & \textbf{83.81} & \textbf{3.50} & \textbf{5.50} & 2.68 & 4.48 & \textbf{9.33} \\
\bottomrule
\end{tabular*}

\end{table}

\noindent\textbf{Filtering and motion speed.}
One Euro filtering operates on the same DARK-decoded~\cite{zhang2020dark} spatial-baseline coordinates in original RGB pixels. The filter uses minimum cutoff 3 Hz, $\beta=0.1$, derivative cutoff 1 Hz, and sampling frequency 30 Hz. Parameters are selected on validation predictions using $\PCKHB(0.5)$ with a $10^{-3}$-weighted normalised MPJAccE penalty, then fixed across test videos and speed strata. In Table~\ref{tab:speed-strata}, SpikePose has the lowest overall MPJVE and MPJAccE, improving on One Euro by 3.0\% and 2.5\%.

For high-speed videos, SpikePose reduces MPJAccE from 9.90 to 9.33 relative to One Euro, with a $\PCKHB(0.5)$ trade-off from 73.13 to 72.88; One Euro has lower low- and mid-speed MPJAccE.


\section{Conclusion}


We presented SpikePose, a lightweight video HPE system that separates spatial and temporal processing: a frame-reset spiking encoder produces per-frame heatmaps with a single internal update, while MAM confines cross-frame reasoning to a compact, aligned heatmap memory outside the backbone. Experiments on MPII and a frozen NTU60 subset show that MAM substantially improves motion consistency at negligible additional cost, and that SpikePose narrows the accuracy gap with its ANN heatmap-based counterpart while retaining the energy advantages characteristic of spiking computation.

\section*{Acknowledgements}

F. Deligianni acknowledges funding from EPSRC (EP/W01212X/1), the Royal Society (RGS/R2/212199), and the Academy of Medical Sciences (NGR1/1678).

\bibliographystyle{IEEEbib}
\bibliography{refs}

\end{document}